\documentclass[11pt]{article}
\usepackage{emnlp2016}

\def\emnlppaperid{11}
\emnlpfinalcopy

\usepackage{times}
\usepackage{latexsym}
\usepackage[utf8]{inputenc}
\usepackage{amssymb}
\usepackage{tikz}
\usepackage{multirow}
\usepackage{booktabs}
\usepackage{enumitem}
\usepackage{graphicx} 
\graphicspath{ {images/} }

\usepackage{amsmath}
\DeclareMathOperator{\softmax}{softmax}
\DeclareMathOperator{\float}{float}
\DeclareMathOperator*{\argmax}{argmax}

\usepackage{algpseudocode,algorithm,algorithmicx}
\newcommand*\Let[2]{\State #1 $\gets$ #2}
\algrenewcommand\algorithmicrequire{\textbf{Input:}}
\algrenewcommand\algorithmicensure{\textbf{Output:}}

\usepackage{xcolor}
\definecolor{nice-red}{HTML}{E41A1C}
\definecolor{nice-orange}{HTML}{FF7F00}
\definecolor{nice-yellow}{HTML}{FFC020}
\definecolor{nice-green}{HTML}{4DAF4A}
\definecolor{nice-blue}{HTML}{377EB8}
\definecolor{nice-purple}{HTML}{984EA3}

\usepackage[textsize=small]{todonotes}
\usepackage{comment}
\usepackage{enumitem}

\title{Clinical Text Prediction\\with Numerically Grounded Conditional Language Models}
\author{
Georgios P. Spithourakis\\
Department of Computer Science\\
University College London\\
g.spithourakis@cs.ucl.ac.uk
\And Steffen E. Petersen\\
William Harvey Research Institute\\
Queen Mary University of London\\
s.e.petersen@qmul.ac.uk
\AND Sebastian Riedel\\
Department of Computer Science\\
University College London\\
s.riedel@cs.ucl.ac.uk
}

\begin{document}

\maketitle

\begin{abstract}

Assisted text input techniques can save time and effort and improve text quality.
In this paper, we investigate how grounded and conditional extensions to standard neural language models can bring improvements in the tasks of word prediction and completion.
These extensions incorporate a structured knowledge base and numerical values from the text into the context used to predict the next word.
Our automated evaluation on a clinical dataset shows extended models significantly outperform standard models.
Our best system uses both conditioning and grounding, because of their orthogonal benefits.
For word prediction with a list of 5 suggestions, it improves recall from 25.03\% to 71.28\%  and for word completion it improves keystroke savings from 34.35\% to 44.81\%, where theoretical bound for this dataset is 58.78\%.
We also perform a qualitative investigation of how models with lower perplexity occasionally fare better at the tasks. We found that at test time numbers have more influence on the document level than on individual word probabilities.

\end{abstract}

\section{Introduction}
\label{sec:intro}

Text prediction is the task of suggesting the next word, phrase or sentence while the user is typing. It is an assisted data entry function that aims to save time and effort by reducing the number of keystrokes needed and to improve text quality by preventing misspellings, promoting adoption of standard terminologies and allowing for exploration of the vocabulary
~\cite{sevenster2010snomed,sevenster2012algorithmic}.

\begin{figure}[!ht]
\centering
\includegraphics[width=0.48\textwidth]{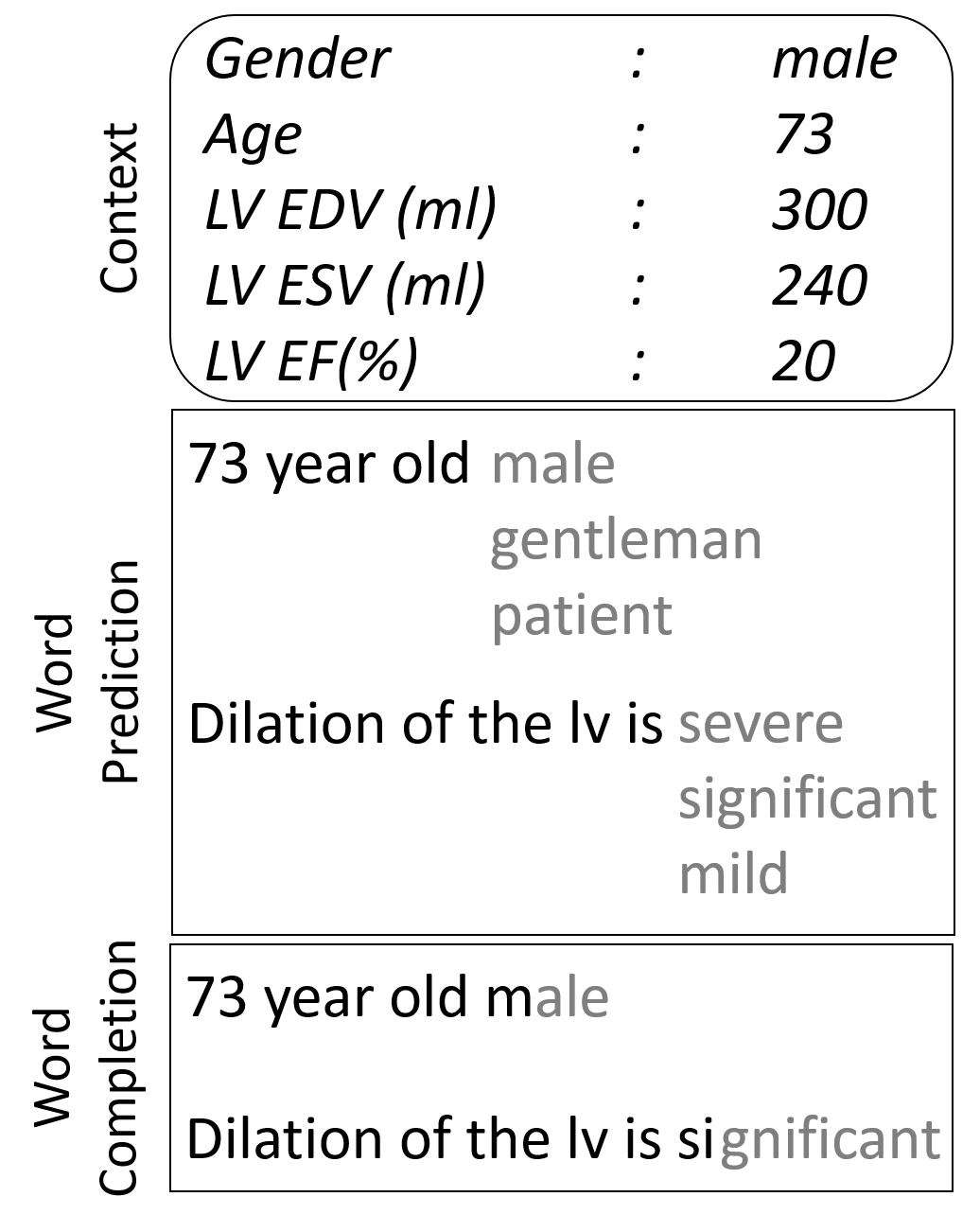}
\caption{Word prediction and completion tasks. A system makes suggestions (in grey) for the next word and to complete a word as it is being typed, respectively. The context is often relevant to the quality of the suggestions.}
\label{fig:task_example}
\end{figure}

Text prediction originated in augmentative and alternative communication (AAC) to increase text generation rates for people with motor or speech impairments~\cite{beukelman2005augmentative}. Its scope has been extended to a gamut of applications, such as 
data entry in mobile devices~\cite{dunlop2000predictive},
interactive machine translation~\cite{foster2002user},
search term auto-completion~\cite{bast2006type} and
assisted clinical report compilation ~\cite{eng2004informatics,cannataro2012knowledge}.


In this paper, we explore the tasks of word prediction, where a system displays a list of suggestions for the next word before the user starts typing it, and word completion, where the system suggests a single possible completion for the word, while the user is typing its characters. The former task is relevant when the user has not yet made a firm decision about the intended word, thus any suggestions can have a great impact in the content of the final document. In the latter case, the user is thinking of a particular word that they want to input and the system's goal is to help them complete the word as quickly as possible. Figure~\ref{fig:task_example} shows examples for both tasks.

Often, the user's goal is to compose a document describing a particular situation, e.g. a clinical report about a patient's condition. An intelligent predictive system should be able to account for such contextual information in order to improve the quality of its suggestions. Challenges to modelling structured contexts include mixed types of values for the different fields an schema inconsistencies across the entries of the structure. We address these issues by employing numerically grounded conditional language models~\cite{anonymous2016gconditional}.

The contribution of this work is twofold. First, we show that conditional and numerically grounded models can achieve significant improvements over standard language models in the tasks of word prediction and completion. Our best model with a list of 5 suggestions raises recall from 25.03\% to 71.28\% and keystroke savings from 34.35\% to 44.81\%.
Second, we investigate in depth the behaviour of such models and their sensitivity to the numerical values in the text. We find that the grounded probability for the whole document is more sensitive to numerical configurations than the probabilities of individual words.






\section{Related Work}
\label{sec:relatedwork}

There have been several applications of text prediction systems in the clinical domain. Word completion has been a feature of discharge summary~\cite{chen2012design}, brain MRI report~\cite{cannataro2012knowledge} and radiology report~\cite{eng2004informatics} compilation systems.
Aiming towards clinical document standardisation, \newcite{tavast2012dynamic} adopted the ICD-10 medical classification codes as a lexical resource and \newcite{lin2014comparison} built a semi-automatic annotation tool to generate entry-level interoperable clinical documents.

~\newcite{hua2014text} reported 13.0\% time reduction and 3.9\% increase of response accuracy in a data entry task.
~\newcite{gong2016leveraging} found a performance of 87.1\% for keystroke savings,
a 70.5\% increase in text generation rate,
a 34.1\% increase in reporting comprehensiveness 
and a 14.5\% reduction in non-adherence to fields when reporting on patient safety event.
In non-clinical applications, a survey of text prediction systems~\cite{garay2006text} reports keystroke savings ranging from 29\% to 56\%.


The context provided to the predictive system can have a significant effect on its performance.
~\newcite{fazly2003testing} and ~\newcite{van2008efficient} obtained significantly better results for word completion by considering not only the prefix of the current word but also previous words and characters, respectively.
~\newcite{wandmacher2008methods} explored methods to integrate n-gram language models with semantic information and ~\newcite{trnka2008adaptive} used topic-adapted language models for word prediction.
More recently, ~\newcite{ghosh2016contextual} incorporated sentence topics as contextual features into a neural language model and reported perplexity improvements in a word prediction task.
None of these systems considers structured background information or numerical values from the text as additional context.


The motivation to include this information as context to text prediction system is based on the importance of numerical quantities to textual entailment systems~\cite{RoyVieiraEtAl2015,sammons2010ask,maccartney2008modeling,DeRaffertyEtAl2008}.
In medical communications, sole use of verbal specifications (e.g. adjectives and adverbs) has been associated with less precise understanding of frequencies~\cite{nakao1983numbers} and probabilities~\cite{timmermans1994roles}.
A combination of structured data and free text is deemed more suitable for communicating clinical information~\cite{lovis2000power}.


Language models have been an integral part of text prediction systems~\cite{bickel2005predicting,wandmacher2008methods,trnka2008adaptive,ghosh2016contextual}.
Several tasks call for generative language models that have been conditioned on various contexts, e.g.
foreign language text for machine translation~\cite{cho2014learning},
images~\cite{vinyals2015show,donahue2015long} and 
videos~\cite{yao2015describing} for captioning, etc.
Grounded language models represent the relationship between words and the non-linguistic context they refer to. Grounding can help learn better representations for the atoms of language and their interactions.
Previous work grounds language on 
vision~\cite{bruni2014multimodal,silberer-lapata:2014:P14-1},
audio~\cite{kiela-clark:2015:EMNLPa},
video~\cite{fleischman2008grounded}
and the olfactory perception~\cite{kiela-bulat-clark:2015:ACL-IJCNLP}.
\newcite{anonymous2016gconditional} use numerically grounded language models and language models conditioned on a lexicalised knowledge base for the tasks of semantic error detection and correction. We directly use their models to perform word prediction and completion.




\section{Methodology}
\label{sec:models}

In this section we present a solution to the word prediction and completion tasks (Subsection ~\ref{sec:word_pred_and_compl}).
Then, we discuss how language models, which can be grounded on numeric quantities mentioned in the text and/or conditioned on external context can be used in our framework (Subsection~\ref{sec:LMs}).
Finally, we describe our automated evaluation process and various evaluation metrics for the two tasks (Subsection~\ref{sec:evaluation}).

\subsection{Word prediction and completion}\label{sec:word_pred_and_compl}

Let $\{w_1, ..., w_T\}$ denote a document, where $w_t$ is the word at position $t$.
Documents are often associated with external context that can be structured (e.g. a knowledge base) or unstructured (e.g. other documents).
Let's consider the case where our context is a knowledge base ($KB$), that is a set of tuples of the form $<attribute, value>$, where attributes are defined by the KB schema.
Different attributes might take values from different domains, e.g. strings, binary values, real numbers etc., and some of the values might be missing.

In the word prediction task, the system presents a ranked list of suggestions for the next word to the user, before the user starts typing. The user can consult this list to explore the vocabulary and guide their decision for the next word to write. The ranking of the items in the list is important, with more strongly endorsed words appearing higher up. Too many displayed options can slow down skilled users~\cite{langlais2002trans}, therefore the list should not be too long.

Typically, a language model is used to estimate the probability of the next word $w_t$ given the typed word history $w_1,..,w_{t-1}$ and external context. The N-best list of the words with the highest probability is presented as the suggestions.

Word completion is a more interactive task, where the system makes suggestions to complete the current word as the user types each character.
Here, the user has a clear intention of typing a specific word and the system should help them achieve this as quickly as possible.
A single suggestion is presented and the user can choose to complete the word, typically by typing a special character (e.g. tab).

\begin{algorithm}[t!]
  \caption{Word completion}
  \label{alg:char_pseudo}
  \begin{algorithmic}[1]
        \Require
        $\mathcal{V}$ is set of vocabulary words,
        $scorer$ returns score for word in current position
        \Ensure next word to be written
    \Function{CompleteWord}{$\mathcal{V}$, $scorer$}
        \Let{$prefix$}{`'}
        \Let{$lexicon$}{$\mathcal{V}$}
        \Loop
            \Let{$lexicon$}{\{tokens in $lexicon$ starting with $prefix$\}}
            \Let{$best$}{$\argmax\limits_{token \in lexicon}scorer(token)$}
            \State Display $best$
            \Let{$char$}{read next char}
            \If{$char=TAB$}
                \State \Return{best}   \Comment{Auto-complete}
            \ElsIf{$char=WHITESPACE$}
                \State \Return{$prefix$} \Comment{Next word}
            \Else
                \Let{$prefix$}{$prefix+char$} \Comment{Append}
            \EndIf
        \EndLoop
    \EndFunction
  \end{algorithmic}
\end{algorithm}

Word completion is based on interactive prefix matching against a lexicon, as shown in Algorithm~\ref{alg:char_pseudo}. The algorithm takes as input the set of known vocabulary words and a scoring function that returns the goodness of a word in the current position and context, which again can be the word probability from a language model.
Initialisation sets the prefix to an empty string and the lexicon to the whole vocabulary (lines 2-3). Iteratively, words that do not match with the prefix are removed from the lexicon (line 5), the best word from the lexicon according to the scorer is found and displayed to the user (lines 6-7) and the user can respond with a key (line 8). If the user inputs the special character, the best word is automatically completed (lines 9-10). If the user inputs a whitespace character, the algorithm terminates (11-12). This is the case when no matching word is found in the vocabulary. If any other character is typed, it is appended to the prefix and another iteration begins.

\subsection{Neural language models}\label{sec:LMs}

A language model (LM) estimates the probability of the next token given the previous tokens, i.e. $p(w_t|w_1,...,w_{t-1})$.
Recurrent neural networks (RNNs) have been successfully used for language modelling~\cite{mikolov2010recurrent}.
Let $w_t$ also denote the one-hot representation of the $t$-th token, i.e. $w_t$ is a sparse binary vector with a single element set to 1, whose index uniquely identifies the token among a vocabulary of $V$ known words.
A neural LM uses a matrix, $E_{in} \in \mathbb{R}^{D \times V}$, to derive word embeddings, $e^w_t = E_{in}w_t$, where $D$ is a latent dimension.
A hidden state from the previous time step, $h_{t-1}$, and the current word embedding, $e^w_t$, are sequentially fed to an RNN's recurrence function to produce the current hidden state, $h_t \in \mathbb{R}^D$. The conditional probability of the next word is estimated as $\softmax(E_{out}h_t)$, where $E_{out} \in \mathbb{R}^{V \times D}$ is an output embeddings matrix.

We use two extensions to the baseline neural LM, described in ~\newcite{anonymous2016gconditional}. A language model can be \emph{conditioned} on the external context by using an encoder-decoder framework. The encoder builds a representation of the context, $h_{KB}$, which is then copied to the initial hidden state of the language model (decoder).
To build such a representation for our structured context, we can lexicalise the KB by converting its tuples into textual statements of the form "$attribute : value$", which can then be encoded by an RNN. This approach can incorporate KB tuples flexibly, even when values of some attributes are missing.

The document and lexicalised KB will frequently contain numerical tokens, which are typically associated with high out-of-vocabulary rates. To make the LM more sensitive to such numerical information, we can define the inputs of the RNN's recurrence function at each time step as a concatenation of $e^w_t$ and $e^n_t$, where the latter is a representation of the numeric value of $w_t$.
We set $e^n_t=\float(w_t)$, where $\float(.)$ returns a floating point number from the string of its input or zero, if the conversion fails.
When we train such a model, the representations for the words will be associated with the numerical values that appear in their context. Therefore, this model is numerically \emph{grounded}.

\subsection{Automated evaluation}\label{sec:evaluation}

We run an automated evaluation for both tasks and all systems by simulating a user who types the text character by character. The character stream comes from a dataset of finalised clinical reports. For the word prediction task, we assume that the word from the dataset is the correct word.
For the word completion task, we assume that the user types the special key to autocomplete the word as soon as the correct suggestion becomes available.

In practice, the two tasks can be tackled at the same time, e.g. a list of suggestions based on a language model is shown as the user types and they can choose to complete the prefix with the word on the top of the list. However, we chose to decouple the two functions because of their conceptual differences, which call for different evaluation metrics.

For word prediction, the user has not yet started typing and they might seek guidance in the suggestions of the system for their final decision. A 
vocabulary exploration system will need to have a high recall. To also capture the effect of the length of the suggestions' list, we will report recall at various ranks (\emph{Recall@k}), where the rank corresponds to the list length. Because our automated evaluation considers a single correct word, Recall@1 is numerically identical to Precision@1. We also report the mean reciprocal rank (\emph{MRR}), which is the multiplicative inverse of the rank of the correct word in the suggestions' list. Finally, per token \emph{perplexity} is a common evaluation metric for language models.


For word completion, the main goal of the system should be to reduce input time and effort for the intended word that is being typed by the user. \emph{Keystroke savings} (KS) measures the percentage reduction in keys pressed compared to character-by-character text entry.
Suggestions that are not taken by the user are a source of unnecessary distractions. We define an \emph{unnecessary distractions} (UD) metric as average number of unaccepted character suggestions that the user has to scan before completing a word.

\begin{equation}
KS = \frac{keys_{unaided} - keys_{with\:prediction}}{keys_{unaided}}
\end{equation}

\begin{equation}
UD = \frac{count(suggested, \: not \: accepted)}{count(accepted)}
\end{equation}

~\newcite{bickel2005predicting} note that KS corresponds to a recall metric and UD to a precision metric. Thus, we can use the F1 score (harmonic mean of precision and recall) to summarise both metrics.

\begin{equation}
Precision = \frac{count(accepted)}{count(suggested)}
\end{equation}

\begin{equation}
Recall = \frac{count(accepted)}{count(total \: characters)}
\end{equation}



\section{Data}
\label{sec:data}
\begin{table}
\centering
 \begin{tabular}{r r r r r r} 
 \hline
 & & & \bf train & \bf dev & \bf test \\ [0.5ex]
 \hline
 \multicolumn{3}{r}{\#documents} & 11,158 & 1,625 & 3,220 \\
 \multicolumn{3}{r}{\#KB tuples/doc} & 7.7 & 7.7 & 7.7 \\
 \hline
 
 \parbox[t]{2mm}{\multirow{3}{*}{\rotatebox[origin=c]{90}{\#tokens/}}}
 &
 \parbox[t]{2mm}{\multirow{3}{*}{\rotatebox[origin=c]{90}{doc}}}
 & all & 204.9 & 204.4 & 202.2\\
 & & words & 95.7\% & 95.7\% & 95.7\%\\
 & & numeric & 4.3\% & 4.3\% & 4.3\%\\
 \hline
 \parbox[t]{2mm}{\multirow{3}{*}{\rotatebox[origin=c]{90}{OOV}}}
 &
 \parbox[t]{2mm}{\multirow{3}{*}{\rotatebox[origin=c]{90}{rate}}}
 & all & 5.0\% & 5.1\% & 5.2\%\\
 & & words & 3.4\% & 3.5\% & 3.5\%\\
 & & numeric & 40.4\% & 40.8\% & 41.8\%\\
 \hline
 \multicolumn{3}{r}{\#chars/token} & 4.9 & 4.9 & 4.9 \\
 \hline

 \end{tabular}
 \caption{Statistics for clinical dataset. Counts for non-numeric (\textit{words}) and \textit{numeric} tokens reported as percentage of counts for \textit{all} tokens. Out-of-vocabulary (OOV) rates are for vocabulary of $1000$ most frequent words in the train data.
 }
 \label{tab:document_statistics}
\end{table}


Our dataset comprises 16,003 anonymised clinical records from the London Chest Hospital. Table~\ref{tab:document_statistics} summarises descriptive statistics of the dataset.

Each patient record consists of a text report and accompanying structured KB tuples.
The latter describe metadata about the patient (age and gender) and results of medical tests (e.g. end diastolic and systolic volumes for the left and right ventricles as measured through magnetic resonance imaging).
This information was extracted from the electronic health records held by the hospital and was available to the clinician at the time of the compilation of the report.
In total, the KB describes 20 possible attributes. From these, one is categorical (gender) and the rest are numerical (age is integer and test results are real valued).
On average, $7.7$ tuples are completed per record.

Numeric tokens account for a large part of the vocabulary ($>$40\%) and suffer from high out-of-vocabulary rates ($>$40\%), despite constituting only a small proportion of each sentence (4.3\%).

\section{Results and discussion}
\label{sec:experiments}
In this section we describe the setup of our experiments (Subsection~\ref{sec:results_setup}) and then present and discuss
evaluation results for the word prediction (Subsection~\ref{sec:results_word}) and word completion (Subsection~\ref{sec:results_char}) tasks. Finally, we perform a qualitative evaluation (Subsection~\ref{sec:qualitative_results}).

\subsection{Setup}
\label{sec:results_setup}

\begin{table*}[!t]
\centering
 \begin{tabular}{r l r r r r r r} 
 \hline
 & \textbf{model} & \textbf{PP} & \textbf{MRR} & \textbf{Recall@1} & \textbf{Recall@3} & \textbf{Recall@5} & \textbf{Recall@10}\\ [0.5ex]
 \hline
 \parbox[t]{2mm}{\multirow{4}{*}{\rotatebox[origin=c]{90}{system}}}
 &baseline & 14.96 & 17.19 & 8.36 & 18.38 & 25.03 & 36.66 \\
 &+c & 14.52 & 54.49 & 45.27 & 59.97 & 65.18 & 71.18 \\
 &+g & 9.91 & 31.91 & 21.13 & 35.45 & 43.66 & 53.72 \\
 &+c+g & \textbf{9.39} & \textbf{60.71} & \textbf{51.76} & \textbf{66.36} & \textbf{71.28} & \textbf{77.10}\\
 \hline
 \parbox[t]{2mm}{\multirow{5}{*}{\rotatebox[origin=c]{90}{ablation}}}
 &+c -kb & 16.64 & 52.54 & 43.07 & 57.89 & 63.66 & 70.45 \\
 &+g -v & 13.16 & 56.08 & 46.58 & 61.96 & 67.30 & 73.49 \\
 &+c+g -kb & 10.82 & 58.72 & 49.46 & 64.31 & 69.71 & 75.98 \\
 &+c+g -v & 11.84 & 57.31 & 47.52 & 63.47 & 68.92 & 75.30\\
 &+c+g -kb-v & 11.81 & 56.61 & 46.68 & 62.78 & 68.48 & 74.87 \\  
 \hline
\end{tabular}
\caption{Word-level evaluation results for next word prediction on the test set. Perplexity (PP), mean reciprocal rank (MRR) and Recall at different ranks. Recall@1 is equivalent to Precision@1. Best system values in \textbf{bold}.}
\label{tab:word_results}
\end{table*}
\begin{table}[t!]
\centering
 \begin{tabular}{r l r r r r} 
 \hline
 & \textbf{model} & \textbf{P} & \textbf{UD} & \textbf{KS(R)} & \textbf{F1} \\ [0.5ex]
 \hline
 \parbox[t]{2mm}{\multirow{2}{*}{\rotatebox[origin=c]{90}{bound}}}
 &theoretical & 100.0  & 0.00 & 58.87 & 74.11 \\
 &vocabulary & 100.0  & 0.00 & 54.48 & 70.54 \\
 \hline
 \parbox[t]{2mm}{\multirow{4}{*}{\rotatebox[origin=c]{90}{system}}}
 &baseline & 13.96 & 6.17 & 34.35  & 19.85 \\
 &+c & 24.34 & 3.11 & 43.17 & 31.13 \\
 &+g & 18.60 & 4.38 & 39.31 & 25.25 \\
 &+g+c & \textbf{26.60} & \textbf{2.76} & \textbf{44.81} & \textbf{33.38} \\
 \hline
 \parbox[t]{2mm}{\multirow{5}{*}{\rotatebox[origin=c]{90}{ablation}}}
 &+c -kb & 24.61 & 3.06 & 44.22 & 31.62 \\
 &+g -v & 26.74 & 2.74 & 45.71 & 33.74 \\
 &+c+g -kb & 26.73 & 2.74 & 45.72 & 33.74 \\
 &+c+g -v & 27.01 & 2.70 & 45.86 & 33.99 \\
 &+c+g -kb-v & 26.90 & 2.72 & 45.79 & 33.89 \\
 \hline
\end{tabular}
\caption{Character-level evaluation results for word completion on the test set. Unnecessary distractions (UD) is inversely related to precision (P). Keystroke savings (KS) are equivalent with recall (R). Best system values in \textbf{bold}.}
\label{tab:character_results}
\end{table}


Our \emph{baseline} LM is a single-layer long short-term memory network (LSTM)~\cite{hochreiter1997long} with all latent dimensions (internal matrices, input and output embeddings) set to $D=50$.
We extend this baseline model using the techniques described in Section~\ref{sec:LMs} and derive a model conditional on the KB (\emph{+c}), a model that is numerically grounded (\emph{+g}) and a model that is both conditional and grounded (\emph{+c+g}).
We also experiment with ablations of these models that at test time ignore some source of information. In particular, we run the conditional models without the encoder, which ignores the KB (\emph{-kb}), and the grounded models without the numeric representations, which ignores the magnitudes of the numerical values (\emph{-v}).

The vocabulary contains the $V=1000$ most frequent tokens in the training set.
Out-of-vocabulary tokens are substituted with $<$\textit{num}$>$, if numeric, and $<$\textit{unk}$>$, otherwise.
We note that the numerical representations are extracted before any masking.
Models are trained to minimise a cross-entropy loss,
with $20$ epochs of back-propagation and gradient descent with adaptive learing rates (AdaDelta)~\cite{journals/corr/abs-1212-5701} and minibatch size set to $64$.
Hyperparameters are based on a small search on the development set around values commonly used in the literature.


\subsection{Word prediction}
\label{sec:results_word}

We show our evaluation results on the test set for the word prediction task in Table~\ref{tab:word_results}.
The conditioned model (+c) achieves double the MRR and quadruple the Recall@1 of the baseline model, despite bringing only small improvements in perplexity.
The grounded model (+g) achieves a more significant perplexity improvement (33\%), but smaller gains for MRR and Recall@1 (85\% and 150\% improvement, respectively).
Contrary to intuition, we observe that a model with higher perplexity performs better in a language modelling task.

The grounded conditional model (+c+g) has the best performance among the systems, with about 5 points additive improvement across all evaluation metrics over the second best.
The benefits from conditioning and grounding seem to be orthogonal to one another.

Recall increases with the length of the suggestion list (equivalent to rank). The increase is almost linear for the baseline, but for the grounded conditional it has a decreasing rate. The Recall@5 for the best model is similar to Recall@10 for the second best, thus allowing for halving the suggestions at the same level of recall.

In the test time ablation experiments, all evaluation metrics become slightly worse with the notable exception of the grounded without numerical values (+g-v), for which MRR and recall at all ranks are dramatically increased. Again, we observe that a worse perplexity does not always correlate with decreased performance for the rest of the metrics.


\subsection{Word completion}
\label{sec:results_char}

\begin{table*}[!t]
\centering
 \begin{tabular}{|c| r r | r r r r |} 
 \hline

\textbf{document} & & \textbf{system}: & \textbf{baseline} & \textbf{+c} & \textbf{+g} & \textbf{+c+g}\\
\hline
\multirow{5}{4cm}{
left ventricular function
analysis results end diastolic volume $<$num$>$ ml end systolic volume $<$num$>$ ml stroke volume $<$num$>$ ml ejection fraction $<$num$>$ \% [...] lv systolic function is moderately impaired . non dilated atria. non dilated rv [...] lv is $<$word$>$ dilated. [...]}
&&&\multicolumn{4}{c|}{\textbf{suggestions}} \\
\cline{4-7} 
&\multirow{5}{*}{\rotatebox[origin=c]{90}{\textbf{rank}}}
& 1 & non & normal & normal & preserved \\
&& 2 & normal & preserved & dilated & normal \\
&& 3 & dilated & non & not & dilated \\
&& 4 & preserved & good & preserved & not \\
&& 5 & not & mild & non & with \\
\cline{4-7} 
&&&\multicolumn{4}{c|}{\textbf{ranks}} \\
\cline{4-7} 
&\multirow{6}{*}{\rotatebox[origin=c]{90}{\textbf{suggestion}}}
& non-dilated & 10 & 11 & 8 & 13 \\
&& dilated & 3 & 8 & 2 & 3 \\
&& non & 1 & 3 & 5 & 7 \\
&& moderately & 41 & 33 & 37 & 36 \\
&& mildly & 6 & 6 & 7 & 6 \\
&& severely & 29 & 23 & 28 & 27 \\
\hline
 
 \end{tabular}
 \caption{Word prediction for sample document from the development set. Top-5 suggestion lists for $<$word$>$ (original document has ``non'') and ranks for interesting terms from the complete lists of different systems.}
 \label{tab:suggestion_list}
\end{table*}

\begin{table}[t!]
\centering
 \begin{tabular}{r r | r r r r}
&& \multicolumn{4}{c}{numerical configuration} \\
& & non & mild & severe \\
\hline
\multirow{3}{*}{\rotatebox[origin=c]{90}{$<$word$>$}}
& non & \textbf{\textit{85.83}} & \textbf{50.45} & 26.81 \\
& mildly & 11.99 & \textit{36.27} & \textbf{46.46} \\
& severely & 2.18 & 13.28 & \textit{26.73} \\
 \end{tabular}
 \caption{Document probabilities for different $<$word$>$ choices and different numerical configurations. The probabilities are re-normalised over the three displayed choices. Probabilities for highest scoring word in \textbf{bold} and for correct word in \textit{italics}.}
 \label{tab:doc_probs}
\end{table}

We show our evaluation results on the test set for the word prediction completion in Table~\ref{tab:character_results}.
In order to give some perspective to the results, we also compute upper bounds originally used to frame keystroke savings~\cite{trnka2008evaluating}.
The \emph{theoretical} bound comes from an ideal system that retrieves the correct word after the user inputs the only the first character. The \emph{vocabulary} bound is similar but only makes any suggestion if the correct word is in the known vocabulary. We extend these bounds to the rest of the evaluation metrics.

The conditioned model (+c) improves the keystroke savings by 25\% over the baseline, while halving the unnecessary distractions.
The grounded model (+g) achieves smaller improvements over the baseline.
The grounded conditional model (+c+g) again has the best performance among the systems. It yields keystroke savings of 44.81\%, almost halfway to the theoretical bound, and the lowest number of unnecessary distractions.

For this task, the desired behaviour of a system is to increase the keystroke savings without introducing too many unnecessary distractions (as measured by the number of wrongly suggested characters per word). Since the two quantities represent recall and precision measurements, respectively, a trade-off is expected between them~\cite{bickel2005predicting}. Our extended models manage to improve both quantities without trading one for the other.

The theoretical and vocabulary bounds represent ideal systems that always make correct suggestions (UD=0). This translates into very high precision (100\%) and F1 values ($>$70\%) that purely represent upper bounds on these performance metrics. For reference, a system with the same keystroke savings as the theoretical bound (58.87\%) and a single unnecessary character per word (UD=1) would achieve precision of 50\% and an F1 score of 54.07\%.

In the test time ablation experiments, all evaluation metrics have slightly better results than their corresponding system. In fact, some models perform similarly to the best system, if not marginally better.

\begin{figure*}[!t]
\centering
\includegraphics[width=0.9\textwidth]{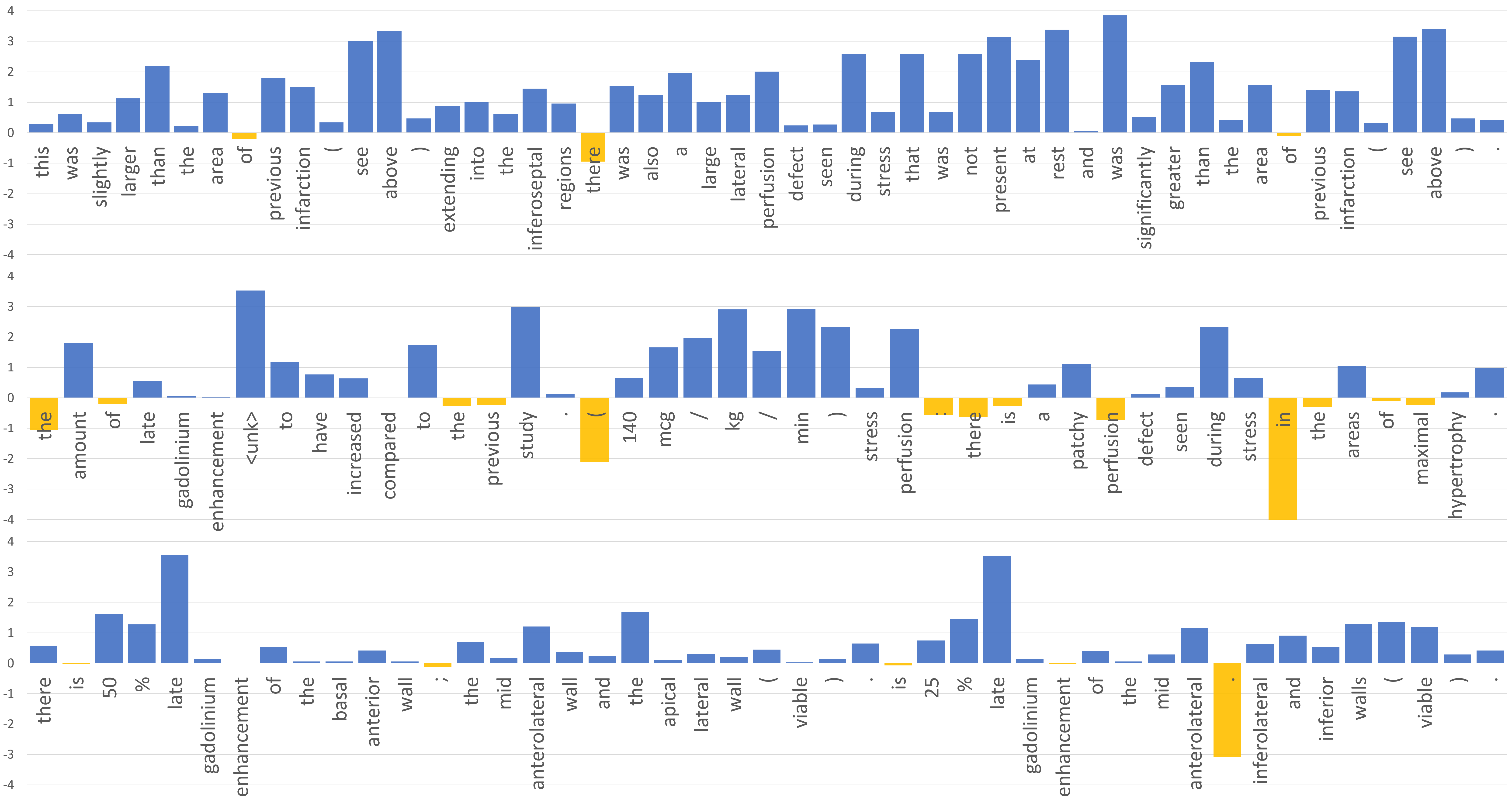}
\caption{Word likelihood ratios (grounded conditional to baseline) for sample sentences from the development set.}
\label{fig:ppmi}
\end{figure*}


\subsection{Qualitative results}
\label{sec:qualitative_results}

The previous results revealed two unexpected situations. First, we observed that occasionally a model with worse perplexity fares better at word prediction, which is a language modelling task. Second, we observed that occasionally a run time ablation of a conditional or grounded model outperforms its system counterpart. We carried out qualitative experiments in order to investigate these scenarios.

We selected a document from the development set and identified a word of interest and numeric values that can influence the user's choice for that word.
In Table~\ref{tab:suggestion_list}, we show the selected document and the 5 top suggestions for the word by different systems. The systems do not have access to tokens from $<$word$>$ onwards.
We also show the ranks for several other semantically relevant choices that appear deeper in the suggestion list. Grounding and conditioning change the order in which the suggestions appear.

We proceeded to substitute the numeric values to more representative configurations that would each favour a particular word choice from the set \{``non'',  ``mildly'',  ``severely''\}. We found that changing the values does not have a significant effect to the suggestion probabilities and causes no reordering of the items in the lists shown in Table~\ref{tab:suggestion_list}. This is in agreement with our previous results for test time ablations and can be attributed to the fact that many more parameters have been used to model words than numerical values. Thus, the systems rely less on numerical information at test time, even though at training time it helps to improve the language models.

Next, for the different numeric configurations we set $<$word$>$ to each of the three choices and computed the probability of observing the whole document under the grounded model. This is done by multiplying together the probabilities for all individual words. Table~\ref{tab:doc_probs} shows the resulting document probabilities, re-normalised over the three choices. We observe that the system has a stronger preference to ``non'', which happens to be the majority class in the training data. In contrast to word probabilities, document probabilities are influenced by the numerical configuration.

The reason for this difference in sensitivities is that the tiny changes in individual word probabilities accumulate multiplicatively to bring on significant changes in the document probability. Additionally, selecting a particular word influences the probabilities of the following words differently, depending on the numerical configuration. This also explains the observed differences between the perplexity of ablated systems, which accumulates small changes over the whole corpus, and the rest of the metrics, which only depend on per word suggestions. Our training objective, cross-entropy, is directly related to perplexity. Through this, numerical values seem to mediate at training time to learn a better language model.

Finally, we directly compare the word probabilities from different systems on several documents from the development set. In Figure~\ref{fig:ppmi} we plot the word likelihood ratio of the grounded conditional to baseline language models for three sentences.
We can interpret the values on the vertical axis as how many times the word is more likely under the extended model versus the baseline.
The probability of most words was increased, even at longer distances from numbers (first example). This is reflected in the improved perplexity of the language model. Words and contingent spans directly associated with numbers, such as units of measurement and certain symbols, also receive a boost (second example). Finally, the system would often recognise and penalise mistakes because of their unexpectedness (dot instead of a comma in the last example).

\section{Conclusion}
\label{sec:conclusion}
In this paper we showed how numerically grounded language models conditioned on an external knowledge base can be used in the tasks of word prediction and completion.
Our experiments on a clinical dataset showed that the two extensions to standard language models have complimentary benefits. Our best model uses a combination of conditioning and grounding to improve recall from 25.03\% to 71.28\% for the word prediction task. In the word completion task, it improves keystroke savings from 34.35\% to 44.81\%, where the upper theoretical bound is 58.78\% for this dataset.
We found that perplexity does not always correlate with system performance in the two downstream tasks.
Our ablation experiments and qualitative investigations showed that at test time numbers have more influence on the document level than on individual word probabilities.

Our approach did not rely on ontologies or fine grained data linkage.
Such additional information might lead to further improvements, but would limit the ability of our models to generalise in new settings.
While our automated evaluation showed that our extended system achieves notable improvements in keystroke savings, a case study would be required to measure the acceptance of such a system and its impact on clinical documentation processes and patient care.
In the past, deployment of text prediction systems in clinical settings has lead to measurable gains in productivity~\cite{hua2014text,gong2016leveraging}.

In the future, we will investigate alternative ways to encode numerical information, in an attempt to improve the utilisation of numerical values at test time. We will also experiment with multitask objectives that consider numerical targets.

\ifemnlpfinal
\section*{Acknowledgments}
The authors would like to thank the anonymous reviewers. This research was supported by the Farr Institute of Health Informatics Research and an Allen Distinguished Investigator award.
\fi

\bibliographystyle{emnlp2016}
\bibliography{bibref}

\end{document}